\documentclass{article}

\usepackage[preprint,nonatbib]{neurips_2025}

\usepackage[utf8]{inputenc} 
\usepackage[T1]{fontenc}    
\usepackage{hyperref}       
\usepackage{url}            
\usepackage{booktabs}       
\usepackage{amsfonts}       
\usepackage{nicefrac}       
\usepackage{microtype}      
\usepackage{xcolor}         
\usepackage{enumitem} 
\usepackage{graphicx}
\usepackage{amsmath}
\usepackage{pifont}
\usepackage{capt-of}
\usepackage{wrapfig}
\usepackage{comment}
\usepackage{multirow}
\usepackage{wrapfig}
\usepackage{lipsum} 
\usepackage{float}
\usepackage{graphicx}

\definecolor{navyblue}{rgb}{0.0, 0.0, 0.5}
\definecolor{citecolor}{HTML}{962C37}
\definecolor{linkcolor}{HTML}{ED1C24}
\hypersetup{
colorlinks=true,
linkcolor=linkcolor,
citecolor=citecolor,
filecolor=navyblue,
urlcolor=navyblue}

\definecolor{BrickRed}{RGB}{203, 65, 84}
\definecolor{ForestGreen}{RGB}{34, 139, 34}
\definecolor{RoyalBlue}{RGB}{65, 105, 225} 
\definecolor{RoyalPurple}{RGB}{120, 81, 169}
\definecolor{OrangeBrown}{RGB}{205, 102, 29}

\newcommand{\cmark}{\textcolor{green!70!black}{\checkmark}}  
\newcommand{\xmark}{\textcolor{red}{\(\times\)}}

\newcommand{\xjqi}[1]{{\color{magenta}{{[\textbf{xjqi}: #1]}}}}
\newcommand{\embraceThreeK}{\mbox{\textcolor{BrickRed}{Emb}\textcolor{ForestGreen}{R}\textcolor{RoyalBlue}{A}\textcolor{RoyalPurple}{C}\textcolor{OrangeBrown}{E}-3K}}

\title{
\textbf{%
\embraceThreeK : %
\textcolor{BrickRed}{Emb}odied \textcolor{ForestGreen}{R}easoning and \textcolor{RoyalBlue}{A}ction in \textcolor{RoyalPurple}{C}omplex \textcolor{OrangeBrown}{E}nvironments
}
}

\begin{document}

\maketitle

\vspace{-45pt}
\begin{center}
    \textbf{
    Mingxian Lin\textsuperscript{1*} \quad 
    Wei Huang\textsuperscript{1*} \quad 
    Yitang Li\textsuperscript{2} \quad 
    Chengjie Jiang\textsuperscript{3} \quad 
    Kui Wu\textsuperscript{4} \\[0.5em]
    Fangwei Zhong\textsuperscript{4} \quad 
    Shengju Qian\textsuperscript{3$\ddagger$} \quad  
    Xin Wang\textsuperscript{3} \quad 
    Xiaojuan Qi\textsuperscript{1\dag}}\\[1em]

    \textsuperscript{1}The University of Hong Kong \quad
    \textsuperscript{2}Tsinghua University\\[0.5em]
    \textsuperscript{3}LIGHTSPEED \quad
    \textsuperscript{4}Beijing Normal University\\[1em]

\href{https://mxllc.github.io/EmbRACE-3K/}{\textcolor{magenta}{https://mxllc.github.io/EmbRACE-3K/}}
\end{center}

\begingroup
\renewcommand\thefootnote{}\footnotetext{* Equal Contribution, $\ddagger$ Project Lead, \dag\ Corresponding Author.}
\endgroup

\begin{abstract}

Recent advanced vision-language models (VLMs) have demonstrated strong performance on passive, offline image and video understanding tasks. However, their effectiveness in embodied settings-- which require online interaction and active scene understanding-- remains limited. In such scenarios, an agent perceives the environment from a first-person perspective, with each action dynamically shaping subsequent observations. Even state-of-the-art models such as GPT-4o, Claude 3.5 Sonnet, and Gemini 2.5 Pro struggle in open-environment interactions, exhibiting clear limitations in spatial reasoning and long-horizon planning. These limitations are further emphasized by our empirical analysis of modern VLMs, which reveals consistent failure modes when applied to embodied tasks. To address this gap, we introduce \embraceThreeK{}, a dataset of over \textbf{3,000} language-guided tasks situated in diverse, photorealistic environments constructed using Unreal Engine and the UnrealCV-Zoo framework. The tasks encompass a wide range of embodied challenges, including navigation, object manipulation, and multi-stage goal execution. Each task unfolds as a multi-step trajectory, pairing first-person visual observations with high-level instructions, grounded actions, and natural language rationales that express the agent's intent at every step. This design results in fine-grained, temporally grounded annotations that closely align perception with decision-making. In total, the dataset contains approximately \textbf{26,000} decision steps, each annotated with multimodal context and step-wise reasoning. Using \embraceThreeK{}, we establish a benchmark to evaluate the embodied reasoning capabilities of VLMs such as GPT-4o, Gemini 2.5 Pro, and Qwen2.5-VL-7B, across three key dimensions: Exploration, Dynamic Spatial-Semantic Reasoning, and Multi-stage Goal Execution. In zero-shot settings, all models achieve success rates below 20\%, underscoring the challenge posed by our benchmark and the current limitations of VLMs in interactive environments. To demonstrate the utility of \embraceThreeK{}, we further fine-tune Qwen2.5-VL-7B using supervised learning followed by reinforcement learning. This approach yields substantial improvements across all three challenge categories, highlighting the dataset's effectiveness in enabling the development of embodied reasoning capabilities. 

\end{abstract}

\section{Introduction}
Recent advances in vision-language models (VLMs) have led to strong performance across a wide range of offline, passive understanding tasks, including image captioning, video summarization, and visual question answering. State-of-the-art models such as GPT-4o~\cite{openai2023gpt4}, Gemini 2.5 Pro~\cite{comanici2025gemini, deepmind2024gemini15}, Claude-3.5-Sonnet~\cite{anthropic2024claude35}, and Qwen2.5-VL~\cite{bai2025qwen2} demonstrate impressive capabilities in aligning visual and linguistic information, particularly when operating on pre-recorded image or vision sequences in static, non-interactive settings. 

\begin{figure*}[!t]
\centerline{\includegraphics[width=1\textwidth]{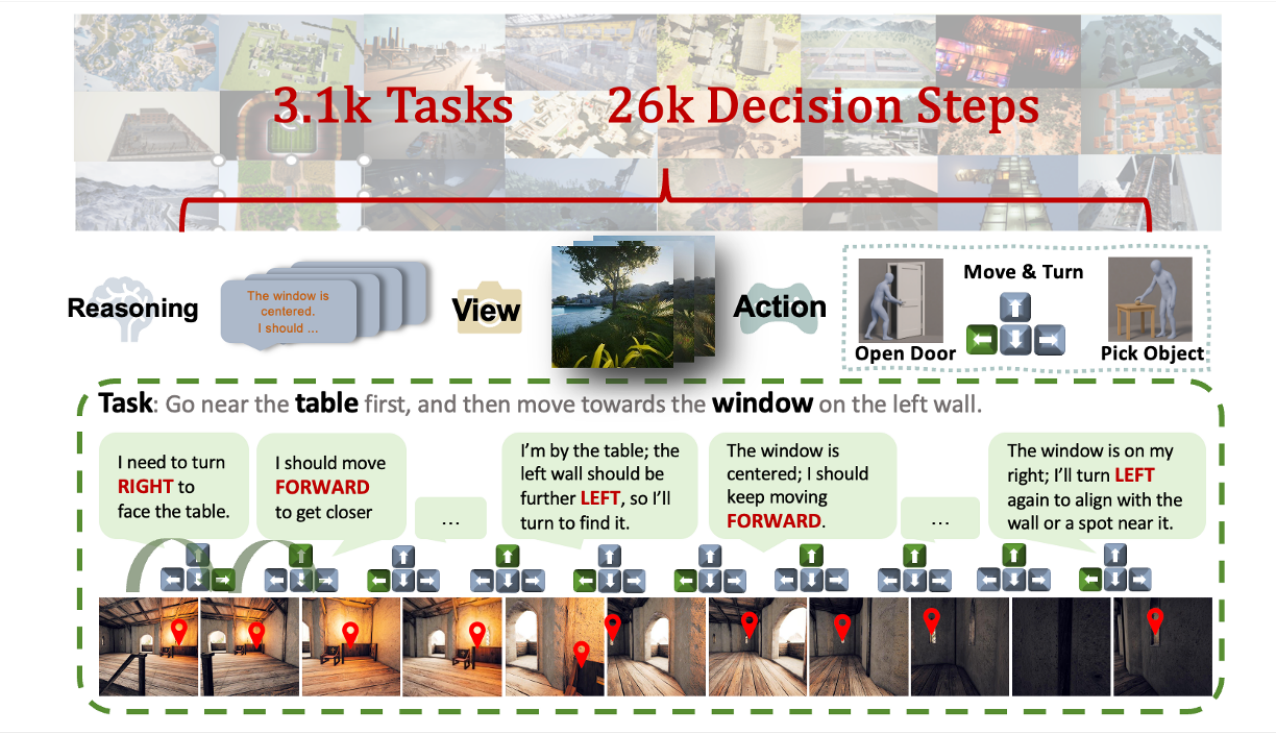}}
\caption{\embraceThreeK{} dataset. The dataset contains over 3k language-guided tasks and 26k decision steps set in diverse, photorealistic environments. Each task involves high-level natural language instructions, grounded actions (e.g., move, turn, look, interact), egocentric visual observations, and step-wise reasoning. Agents interpret visual inputs and follow instructions to execute multi-step decision trajectories, with each step annotated with natural language rationales—forming a coherent and interpretable decision process over a long horizon that involves spatial reasoning.}
\label{fig:teaser}
\vspace{-5pt}
\end{figure*}

However, these models often fall short when applied to \textbf{embodied tasks}, where an agent must actively perceive, reason, and act within an interactive environment~\cite{cheng2024navila,jiao2025free}. Unlike passive vision benchmarks, embodied scenarios involve a \textbf{closed-loop perception-action cycle}: what the agent sees next is determined by the actions it takes now. A single turn or misstep can dramatically alter subsequent observations. In this dynamic, egocentric setting, agents must follow high-level instructions, adapt to constantly shifting visual inputs, and make temporally coherent decisions under partial observability. This tight coupling between perception and action introduces challenges that go far beyond object recognition or static scene understanding, requiring reasoning over how decisions shape future inputs-- posing a fundamentally different learning problem.

Despite this fundamental shift, existing VLMs are often deployed in embodied settings without structural adaptation. In practice, they process short video clips or image sequences as static input, ignoring the dynamic nature of egocentric interaction. This results in a training-deployment mismatch. In our preliminary experiments with Qwen2.5-VL and GPT-4o in simulated embodied environments, we observed consistent failure patterns: the models tend to overfit to immediate visual cues, fail to adjust spatial reasoning as the viewpoint changes, and struggle to maintain attention on objects that briefly exit the field of view. These issues underscore the limitations of passive pretraining when applied to sequential, interactive decision-making tasks. 

To address this gap, we introduce \embraceThreeK{}: \textcolor{BrickRed}{Emb}odied \textcolor{ForestGreen}{R}easoning and \textcolor{RoyalBlue}{A}ction in \textcolor{RoyalPurple}{C}omplex \textcolor{OrangeBrown}{E}nvironments (shown in Figure~\ref{fig:teaser}). This dataset comprises over 3,000 language-guided tasks collected in diverse, photorealistic Unreal Engine environments and controlled via the UnrealCV-Zoo framework. Each task unfolds as a multi-step trajectory in which the agent receives a high-level instruction and interacts with the environment through vision and action. In total, \embraceThreeK{} includes approximately 26,000 decision steps, each annotated with egocentric visual observations, the selected action, and a natural language ``thinking'' rationale that explains the agent’s intent.

Unlike prior datasets, \embraceThreeK{} is built to capture the \textbf{causal structure} of embodied interaction by explicitly modeling how decisions affect perception, and how perception guides subsequent reasoning. It provides fine-grained, step-level annotations that align not only with what the agent observes and does, but also with \textit{why} it acts. These annotations include intermediate reasoning steps, enabling models to learn perception-conditioned decision making rather than relying solely on end-to-end action prediction. Crucially, \embraceThreeK{} supports \textbf{online}, \textbf{closed-loop} interaction, where each action taken by the agent dynamically changes the environment and influences future observations. This setup allows for realistic, temporally extended evaluation under partial observability, supporting spatial-semantic consistency and long-horizon goal pursuit. Compared to prior work such as Octopus~\cite{yang2024octopus}, which formulates reasoning at the level of code generation without step-wise visual grounding, or datasets like ALFRED~\cite{shridhar2020alfred} that rely on pre-defined trajectories, \embraceThreeK{} enables fine-grained, multimodal alignment between perception, language, reasoning, and action. This makes it a strong foundation for evaluating and training embodied agents that not only act, but also interpret, reason, and adapt over time in photo-realistic simulated environments.

By shifting the focus from passive visual comprehension to \textbf{instruction-guided, step-wise reasoning}, \embraceThreeK{} enables the training and evaluation of vision-language agents in goal-oriented embodied scenarios. First, we use \embraceThreeK{} to establish a benchmark targeting three core embodied capabilities: \textbf{Exploration}, \textbf{Dynamic Spatial-Semantic Reasoning}, and \textbf{Multi-stage Goal Execution}. In zero-shot evaluations, state-of-the-art models-- GPT-4o, Gemini 2.5 Pro, and Qwen2.5-VL-7B—achieve success rates below 20\% on all three tasks, revealing a significant gap between current VLM capabilities and the demands of embodied reasoning.
Next, we fine-tune Qwen2.5-VL-7B on \embraceThreeK{} using a two-stage approach: supervised fine-tuning (SFT) followed by reinforcement learning (RL). This significantly boosts the model’s performance across all tasks, resulting in higher success rates and lower goal distance error (GDE) compared to GPT-4o and Gemini 2.5 Pro. These results validate the effectiveness of \embraceThreeK{} in enabling VLMs to acquire embodied reasoning capabilities.
Moreover, we observe that models trained with SFT alone perform well on in-domain tasks, but suffer marked performance degradation on out-of-domain scenarios. This limited generalization highlights the importance of reinforcement-based adaptation for improving robustness in unfamiliar environments. It also underscores the pressing need for high-quality, interaction-centric datasets to support training in such settings-- an area where \embraceThreeK{} is particularly well-positioned to contribute.

\section{Related Work}
\vspace{-1pt}
\paragraph{Vision-Language Models} Recent advancements in vision-language models (VLMs) have demonstrated remarkable progress in both architectural innovations and performance. For example, GPT-4o~\cite{gpt4o} has achieved significant improvements in visual understanding by leveraging enhanced reasoning capabilities. Similarly, Gemini Pro 1.5~\cite{team2024gemini} has extended the context length to an unprecedented 1 million tokens, positioning itself as a leader on long video benchmarks~\cite{fu2024video}. The open-source VLM ecosystem has also seen substantial growth, driven by improvements in model architecture and training methodologies. Notable contributions include state-of-the-art models such as InternVL3~\cite{zhu2025internvl3}, Qwen2.5-VL~\cite{yang2024qwen2}, LLaVA-OneVision~\cite{li2024llava}, and Llama3.2-Vision~\cite{llama3.2}. These advancements have significantly narrowed the performance gap between open-source and proprietary VLMs, with many open-source models now achieving competitive results.

\paragraph{Embodied Tasks for VLM}
Recent efforts in embodied vision-language modeling have targeted diverse forms of perception, control, and supervision. 
Octopus~\cite{yang2024octopus} leverages code synthesis to bridge language and action planning, incorporating environmental feedback in simulation. 
SayCan~\cite{ahn2022can} grounds language in executable actions using affordance-based filtering. 
Ego4D~\cite{grauman2022ego4d} focuses on passive egocentric video understanding without action execution, while TEACh~\cite{padmakumar2022teach} centers on dialog-driven task planning in multi-agent settings. 
Octopi~\cite{yu2024octopi} extends grounded reasoning to the tactile modality.
Despite these advances, many benchmarks lack fine-grained supervision, online interaction, or visual realism. 
\embraceThreeK{} addresses these limitations by providing a fully \textit{step-wise}, \textit{spatio-temporally grounded}, and \textit{closed-loop} evaluation framework built in \textit{photo-realistic} Unreal Engine environments. 
Each decision step is paired not only with egocentric observations and grounded actions, but also with explicit \textit{reasoning annotations} that capture the agent's intent and intermediate thinking process.
This enables more interpretable and diagnostic evaluation of vision-language models in long-horizon embodied tasks.

\begin{table*}[t]
\centering
\caption{Comparison of Embodied Benchmarks for VLM Evaluation across Multiple Dimensions}
\resizebox{\textwidth}{!}{
\begin{tabular}{l|ccccccc}
\toprule
\textbf{Benchmark} 
& \textbf{Input Modality} 
& \textbf{Scene} 
& \textbf{Fidelity Level} 
& \textbf{Step-wise} 
& \textbf{Spatio-Temporal Aware} 
& \textbf{Online} 
& \textbf{Closed-loop} \\
\midrule
ALFRED~\cite{shridhar2020alfred}           & Language\&Vision & Indoor & Photo-Realistic     & \cmark     & \xmark & \xmark   & \cmark     \\
Octopus~\cite{yang2024octopus}          & Language\&Vision & Indoor\&Outdoor & Game-based  & \xmark     & \cmark & \cmark     & \cmark     \\
HabitatNav~\cite{habitatchallenge2023}       & Vision     & Indoor & Real-world     & \cmark     & \cmark     & \cmark      & \xmark     \\
MindCube~\cite{yin2025spatial}            & Vision     & Indoor       & Real-world            & \xmark     & \cmark     & \xmark     & \xmark     \\
V-IRL~\cite{yang2024v}            & Language\&Vision     & Outdoor        & Real-world          & \cmark     & \cmark     & \cmark    & \xmark     \\
VSI~\cite{yang2025thinking}            & Vision     & Indoor        & Real-world          & \xmark     & \cmark     & \xmark     & \xmark     \\
MCU~\cite{lin2023mcu} & Language\&Vision  & Indoor\&Outdoor & Game-based  & \cmark     & \xmark  & \xmark      & \cmark    \\
\midrule
\embraceThreeK{}    & Language\&Vision  & Indoor\&Outdoor   & Photo-Realistic  & \cmark    & \cmark & \cmark & \cmark \\
\bottomrule
\end{tabular}
}
\label{tab:embodied-benchmark-comparison}
\end{table*}

\section{Pilot Study}
\vspace{-1pt}
We conducted a set of preliminary evaluations using GPT-4o and Qwen2.5-VL in simulated embodied environments. Despite architectural differences, both models exhibited similar failure patterns when tasked with step-wise, instruction-driven tasks. These observations highlight fundamental limitations of current video-trained VLMs in embodied settings.

 \textbf{(i) Short-sighted exploration.} The models tend to fixate on immediate visual cues, lacking the ability to plan toward long-term goals. In the task \texttt{"Locate and approach the red car"}, for instance, agents briefly glance left, find no immediate evidence of the target, then glance right with the same result, and promptly proceed forward without conducting a broader search. This behavior suggests that each movement is selected based on local visual feedback rather than an integrated exploration strategy. In \texttt{"Find the plant near the shelf"}, agents abandon shelf traversal after only a few head turns, missing nearby objects outside their immediate field of view.  
This limitation can be traced to how VLMs are typically trained. In conventional video datasets, the model passively receives visual input and learns to summarize or answer questions based on full-sequence observations. There is no need to actively decide where to look or how to explore, so these models never acquire the capacity for information-seeking behavior.

\textbf{(ii) Dynamic spatial-semantic drift.} The interpretation of spatial relations becomes unstable as the agent moves, due to a lack of egocentric pose awareness. In tasks like \texttt{"Approach the second trash can"} or \texttt{"Go to the white house in front"}, agents initially respond correctly to spatial cues but fail to adapt as their viewpoint changes. Ordinal and directional terms such as "second" and "front" become detached from the agent's current orientation, leading to consistent semantic misalignment.  
This issue arises because most VLMs are trained on static or loosely time-linked image-text data. Even in video-based pretraining, spatial reasoning is often limited to temporal QA, captioning, or event ordering, where egocentric position and spatial frame shifts are rarely encoded. As a result, these models rely on static geometric assumptions that do not update as the agent moves, causing a gradual drift in language grounding.

\textbf{(iii) Target forgetting.} Models often fail to retain intent beyond the current frame, leading to target loss when objects briefly leave the field of view. For instance, in \texttt{"Go near the red car"}, a temporary disappearance of the car results in its permanent omission. Similarly, in multi-stage instructions like \texttt{"First reach the trash can, then go to the red car"}, the agent frequently completes the first task but fails to recall the second goal after unrelated actions. This issue stems from the pretraining objectives of most video-language datasets, which emphasize frame-level recognition, counting, or sequence-level QA, rather than persistent target awareness over time. As a result, sudden appearances or disappearances of objects are not treated as meaningful, leaving the model unable to track unseen but relevant entities.

These behaviors expose critical gaps in current VLM capabilities and motivate the design of \embraceThreeK{} to support better spatial reasoning, goal continuity, and instruction-grounded action planning.

\section{Data Collection and Benchmark Construction}

\begin{figure*}[!t]
\centerline{\includegraphics[width=1\textwidth]{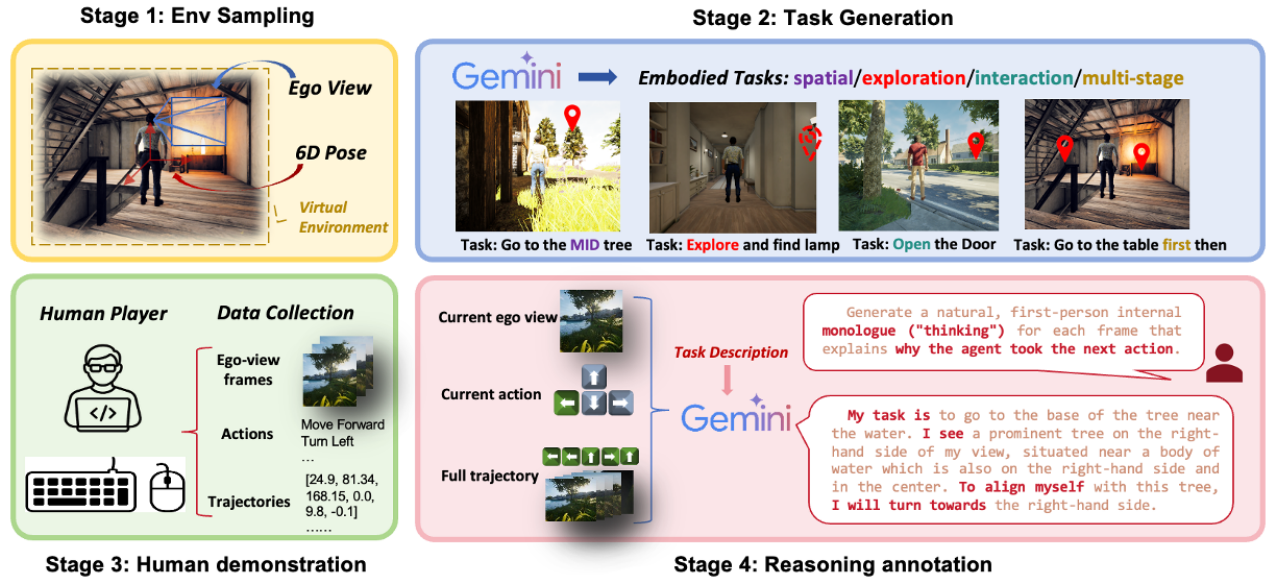}}
\caption{Multi-stage Embodied Task Data Collection Pipeline.
The \embraceThreeK{} dataset is built in four stages: (1) sampling diverse 6-DoF agent poses with ego views in virtual environments, (2) generating grounded task instructions using Gemini, (3) collecting human demonstrations, and (4) annotating each action with step-wise natural language reasoning to explain agent decisions and enhance interpretability.}
\label{fig:data_generation}

\end{figure*}

To support systematic investigation of embodied vision-language reasoning, we construct \embraceThreeK: \textcolor{BrickRed}{Emb}odied \textcolor{ForestGreen}{R}easoning and \textcolor{RoyalBlue}{A}ction in \textcolor{RoyalPurple}{C}omplex \textcolor{OrangeBrown}{E}nvironments. This benchmark is designed to expose the structural mismatch between passively trained VLMs and the demands of step-wise, instruction-driven embodied tasks. Informed by our pilot study, which highlighted issues such as short-sighted action selection, spatial-semantic misalignment, and goal forgetting, we develop a multi-stage data pipeline (shown in Figure.~\ref{fig:data_generation}) that provides fine-grained supervision aligned with closed-loop reasoning.

\vspace{-6pt}
\subsection{Simulation Platform and Environmental Diversity}

All data in \embraceThreeK{} are collected within the UnrealCV-Zoo framework~\cite{unrealzoo}, which extends Unreal Engine with first-person control and low-level API access. From 100 available photorealistic environments, we select 24 diverse maps that span indoor and outdoor settings, varying in object density, spatial topology, lighting, and navigational complexity. This diversity enables robust evaluation of generalization across scene types and task variations.

\subsection{Multi-stage Embodied Task Data Collection}

Our data collection proceeds in four structured stages, designed to capture the full perception-reasoning-action loop required for interactive embodied tasks. This process emphasizes both scene diversity and reasoning complexity to facilitate robust agent training and evaluation.

\paragraph{Stage 1: Environment Sampling and Pose Selection}

We begin by sampling diverse agent poses across selected simulation maps using a hybrid strategy that combines automated scripts and manual inspection. The automated component leverages the Navigation Movement API provided by Unreal Engine to uniformly explore traversable regions. To ensure data quality, sampled positions undergo manual validation to filter out visually trivial locations (e.g., texture-less walls) or physically unreachable areas (e.g., positions blocked by obstacles or un-navigable terrain).
Each selected pose is recorded with full 6-DoF (degrees of freedom) coordinates—including position and orientation as well as the corresponding egocentric RGB image captured from the agent's first-person perspective. 

\paragraph{Stage 2: Task Instruction Generation}

For each selected agent pose, we retrieve object-level metadata within a 1000 meter radius, including the semantic names and spatial positions of nearby objects. This contextual information, along with the egocentric RGB view captured at the pose, is provided to the Gemini 2.5 pro model to generate natural language task instructions. The model is explicitly conditioned on the spatial layout and visual context to ensure semantic grounding, producing instructions that are both plausible and solvable within the local environment.

We also inform the model of the desired task type prior to generation, guiding it to produce tasks aligned with one of five categories identified in our pilot analysis of embodied reasoning challenges:

\noindent
\textit{\textbf{0. Basic:} Target is clearly visible and immediately reachable, requiring minimal reasoning.} \\
\textit{\textbf{1. Exploration:} Target is initially out of view, prompting the agent to perform an active search.} \\
\textit{\textbf{2. Dynamic Spatial-Semantic:} Target is described using relative or ordinal spatial references. }\\
\textit{\textbf{3. Multi-stage:} Task requires completing a series of subgoals in a specific order.}\\
\textit{\textbf{4. Interaction:} Task requires direct manipulation (e.g., open a door, pick or drop an object).}

To ensure quality and diversity, all generated instructions undergo a post-processing stage that includes manual verification and targeted manual authoring. Annotators inspect generated instructions to verify consistency with visual and spatial context, correct ambiguous phrasing, and supplement the dataset with novel, human-authored tasks for underrepresented cases. This hybrid generation-and-curation pipeline ensures both scalability and high-quality alignment with embodied agent capabilities.

\paragraph{Stage 3: Human Demonstration and Trajectory Capture}

Each generated instruction is performed by a human player controlling the agent in real time. We record all egocentric frames, executed actions, and precise pose trajectories. These demonstrations provide high-quality behavioral examples that encode closed-loop dependencies between perception, action context, and intent. The resulting action sequences are typically sparse and efficient, reflecting realistic strategies for exploration and goal completion.

\paragraph{Stage 4: Step-wise Reasoning Annotation}

To improve interpretability and facilitate cognitive supervision, we annotate each step in the demonstration trajectory with natural language explanations of the chosen actions. Unlike traditional chain-of-thought (CoT)~\cite{shao2024visual} methods that focus on isolated frames, our annotations are grounded in the agent’s egocentric perspective and the full task context. Gemini receives the task instructions, complete egocentric views, and the entire action trajectory, enabling holistic reasoning about how each action contributes to the final goal and influences future observations. These explanations capture not only the action taken but also its relevance to the spatial structure, task dynamics, and overall intent. This approach ensures that CoT traces provide decision-level supervision tightly aligned with the perception-action cycle.

\begin{figure*}[!t]
\centerline{\includegraphics[width=1\textwidth]{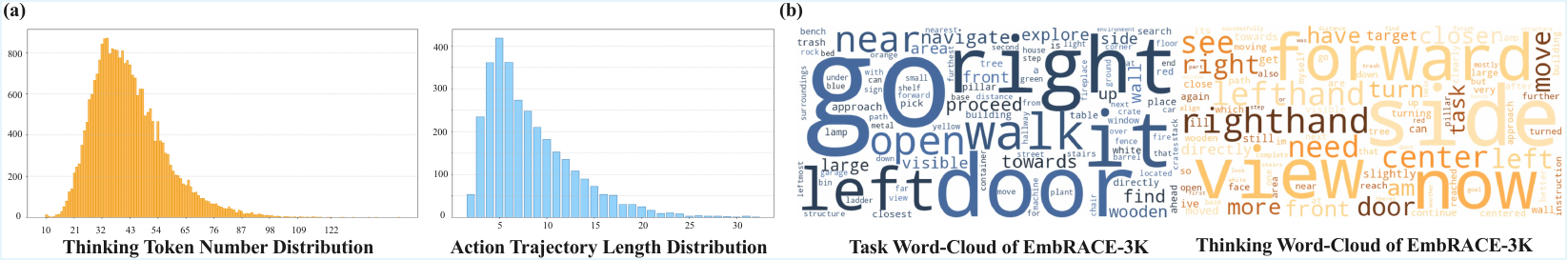}}
\caption{The Token Number and Word-Cloud Distribution of \embraceThreeK{}. 
(a)Token number distribution of reasoning and the length distribution of action trajectories.
(b)The word clouds of task instructions and agent thinking processes in \embraceThreeK{}.
}
\label{fig:data_statistic}
\end{figure*}

\subsection{Data Curation}

\begin{wrapfigure}{r}{0.5\textwidth}
\centering
\includegraphics[width=\linewidth]{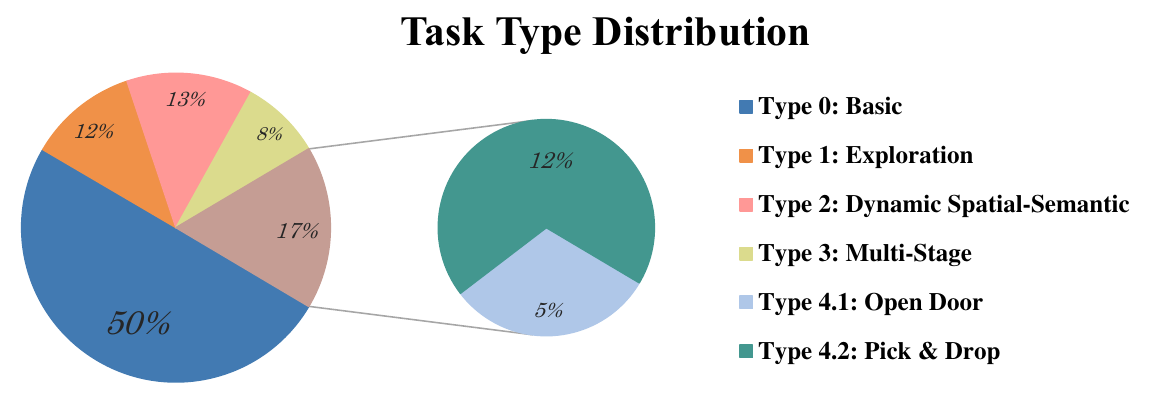}
\vspace{-5pt}
\caption{Distribution of task types.}
\label{fig:data_type_distribution}
\end{wrapfigure}

To ensure high-quality and interpretable data, we apply a series of post-processing and analysis steps to refine the raw collection. First, we filter out trajectories with more than 32 steps to simplify training and evaluation, ensuring consistent sequence length across tasks. Second, we categorize all instructions into five high-level task types based on reasoning demands: Basic, Exploration, Spatial-Relational, Multi-stage, and Interaction. The Interaction class is further subdivided into two subtypes, Open the Door and Pick and Drop the Object, based on UnrealZoo's interaction primitives. The resulting type distribution is shown in Figure~\ref{fig:data_type_distribution}, with Basic tasks accounting for approximately half of the dataset.

In addition to type-based categorization, we conduct a series of corpus-level analyses to characterize the dataset. Figure~\ref{fig:data_statistic}(a) illustrates the distributions of action trajectory lengths and reasoning token counts, showing that most tasks involve under 15 steps and 80 thinking tokens. Figure~\ref{fig:data_statistic}(b) presents word clouds derived from both task instructions and CoT annotations, revealing distinct vocabularies for goal specification and intermediate reasoning.


Finally, we standardize all trajectories into a unified format, including ordered egocentric frames, discrete action sequences, 6-DoF poses, and aligned language fields such as instruction, thinking trace, and step-level justification. Visual content is normalized for resolution and field of view to ensure consistency across samples. These steps ensure that \embraceThreeK{} offers not only scale and diversity, but also the structural coherence required for training and evaluating embodied vision-language models.

\vspace{1pt}


\begin{figure*}[!t]
\centerline{\includegraphics[width=1\textwidth]{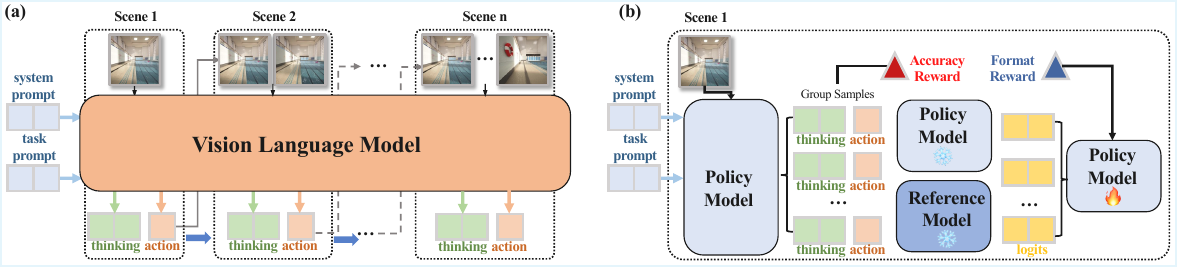}}
\caption{Two-stage Training framework for Embodied Agent on \embraceThreeK{}. (a) SFT training pipeline for agent in open-environment; (b) GRPO training pipeline for agent in open-environment}
\label{fig:training-framework}
\end{figure*}

\section{Reasoning Training Pipeline}
As illustrated in Figure~\ref{fig:training-framework}, we designed two distinct training frameworks within the \embraceThreeK{} dataset to enhance the spatial understanding and action planning capabilities of embodied agents. Specifically, we proposed a supervised fine-tuning (SFT) pipeline based on spatial reasoning, which leverages memory learning to strengthen the agent's reasoning abilities for scenes and actions. Additionally, we developed an exploratory reasoning framework grounded in reinforcement learning, where a rule-based reward function enables the agent to autonomously learn reasoning skills.

\subsection{Supervised Fine-tuning for Reasoning Memory}
\vspace{-2pt}
We utilized Qwen2.5-VL-7B as the foundational models and trained with 2,344 high-quality reasoning trajectories from \embraceThreeK{}, encompassing a total of 10k trainable actions. This training phase was designed to endow the model with enhanced capabilities for understanding new visual scenes and reasoning for action decision-making. As shown in Figure~\ref{fig:training-framework}(a), we directly deployed Qwen2.5-VL-7B into the Llama-Factory~\cite{zheng2024llamafactory} framework for instruction-based SFT through multi-turn dialogues. The supervised training outputs consisted of two key components: the reasoning process enclosed in the \textit{<think></think>} tag and the final action decision enclosed in the \textit{<action></action>} tag. The SFT process was conducted on 8 GPUs.

\subsection{Reinforcement Learning for Reasoning Exploring}
\vspace{-2pt}
Recent advancements~\cite{shao2024deepseekmath,liu2024codemind} suggest that reinforcement learning fine-tuning can lead to breakthroughs in reasoning capabilities across domains such as mathematics and coding. Additionally, reinforcement learning has achieved promising results in multimodal understanding tasks involving images and videos. Building on the observed advancements of the group relative policy optimization (GRPO) algorithm in DeepSeek-R1~\cite{guo2025deepseek} and prior explorations of multimodal reasoning training~\cite{feng2025video, peng2025lmm}, we further adopt the standard GRPO framework, as illustrated in Figure.~\ref{fig:training-framework}(b), to investigate the impact of reinforcement learning on the reasoning abilities of embodied agents using the \embraceThreeK{} dataset. For a given question $q$, the policy model generates a group of candidate responses $\{o_1, o_2,...,o_G\}$ using the previous policy $\pi_{\theta_{old}}$. Each candidate response is associated with a corresponding reward $\{r_1, r_2,...,r_G\}$, which is computed based on rule-based reward functions, such as those evaluating format and accuracy. The updated model $\pi_{\theta}$ is subsequently trained by maximizing the following objective function:

\begin{equation}
\begingroup
\thinmuskip=2mu \medmuskip=2mu \thickmuskip=2mu
\scalebox{0.92}{$
J_{GRPO}(\theta) = \mathbb{E}_{q,\{o_i\}} \left[ \frac{1}{G} \sum_{i=1}^{G} \left(\min\left(\frac{\pi_{\theta}(o_i|q)}{\pi_{\theta_{old}}(o_i|q)}A_i, \mathrm{clip}\left(\frac{\pi_{\theta}(o_i|q)}{\pi_{\theta_{old}}(o_i|q)}, 1-\epsilon, 1+\epsilon\right)A_i\right)- \beta\mathbb{D}_{KL}(\pi_\theta||\pi_{ref})\right)\right]
$}
\label{eq1:grpo}
\notag
\endgroup
\end{equation}

where, $\epsilon$ and $\beta$ are hyperparameters. Given that \embraceThreeK{} contains action trajectories of up to 32 steps in length and corresponding input vision, we set $G = 6$ to normalize the sampled rewards, thereby computing the advantage, $A_i = \frac{r_i - \mathrm{mean}(\{r_1, r_2,...,r_G\})}{\mathrm{std}(\{r_1, r_2,...,r_G\})}$, for updating the model. This approach aims to guide the embodied agent in freely exploring reasoning strategies within open environments. The rule-based supervision incorporates a reward format in the form of \textit{<think></think>} and \textit{<action></action>}, directly evaluating the content of the actions. We conducted GRPO training on the R1V~\cite{chen2025r1v} framework using 8 GPUs.

\section{Experiments of Embodied Agent}
\vspace{-1pt}

\begin{table*}[t]
\centering
\caption{Performance comparison on Type Basic, Exploration and Dynamic Spatial-Semantic tasks.}
\vspace{1mm}
\scriptsize
\renewcommand{\arraystretch}{1}

\resizebox{\textwidth}{!}{
\begin{tabular}{l|ccccc|ccccc|ccccc}
\toprule
\textbf{Method} 
& \multicolumn{5}{c|}{\textbf{Basic}} 
& \multicolumn{5}{c|}{\textbf{Exploration}} 
& \multicolumn{5}{c}{\textbf{Dynamic Spatial-Semantic}} \\
& SR $\uparrow$ & GDE $\downarrow$ & SSPL $\uparrow$ & Steps $\downarrow$ & TR $\downarrow$
& SR $\uparrow$ & GDE $\downarrow$ & SSPL $\uparrow$ & Steps $\downarrow$ & TR $\downarrow$
& SR $\uparrow$ & GDE $\downarrow$ & SSPL $\uparrow$ & Steps $\downarrow$ & TR $\downarrow$ \\
\midrule
\multicolumn{16}{c}{\textbf{In-Domain}} \\
\midrule
GPT-4o                   & {53.6\%}     & {484.7}  & {0.396}     & {17.9}   & {37.1\%}  
                         & {14.3 \%}     & {1178.3}   & {0.086}     & {28.5}   & {75.0\%}  
                         & {62.9\%}     & {374.1}   & {0.521}     & {12.5}   & {25.7\%}\\
Gemini 2.5 Pro           & {76.4\%}     & {232.2}   & {0.649}     & {10.1}   & {13.6\%}  
                         & {39.3 \%}     & {1068.9}   & {0.264}     & {24.3}   & {57.1\%}  
                         & \textbf{71.4\%}     & \textbf{238.1}   & {0.589}     & {13.1}   & {25.7\%}\\
Qwen2.5-VL-origin        & {26.4\%}     & {531.6}   & {0.176}     & {23.4}   & {65.7\%}  
                         & {0.0 \%}     & {991.8}   & {0.000}     & {28.7}   & {85.7\%}  
                         & {14.3\%}     & {527.9}   & {0.079}     & {28.0}   & {80.0\%}\\
Qwen2.5-VL-no-thinking   & {79.3\%}     & {232.4}   & \textbf{0.775}     & \textbf{4.3}   & \textbf{0.7\%}  
                         & {28.6 \%}     & {652.1}   & {0.268}     & \textbf{8.3}   & \textbf{3.6\%}  
                         & {68.6\%}     & {298.5}   & {0.580}     & \textbf{6.0}   & \textbf{0.0\%}\\
Qwen2.5-VL-sft-only      & {72.9\%}     & {237.9}   & {0.647}     & {7.9}   & {4.3\%}  
                         & {\textbf{71.4 \%}}     & {\textbf{279.3}}   & {\textbf{0.594}}     & {15.1}   & {7.1\%}  
                         & {68.6\%}     & {245.0}   & {0.557}     & {10.1}   & {2.9\%}\\
Qwen2.5-VL-sft-rl        & \textbf{81.4\%}     & \textbf{215.7}   & {0.766}     & {6.2}   & {3.6\%} 
                         & {60.7 \%}     & {391.8}   & {0.578}     & {11.9}   & {7.1\%}  
                         & {68.6\%}     & {238.4}   & \textbf{0.612}     & {7.3}   & {2.9\%}\\
\midrule
\multicolumn{16}{c}{\textbf{Out-of-Domain}} \\
\midrule
GPT-4o                   & {20.8\%}     & {1278.6}  & {0.163}     & {20.9}   & {45.4\%}  
                         & {3.6 \%}     & {4017.8}   & {0.011}     & {30.9}   & {90.9\%}  
                         & {10.2\%}     & {2144.2}   & {0.078}     & {26.2}   & {67.8\%}\\
Gemini 2.5 Pro           & {38.0\%}     & {643.8}  & {0.336}     & {12.4}   & {16.2\%}  
                         & {9.1 \%}     & {2166.5}   & {0.077}     & {25.3}   & {60.0\%}  
                         & {20.3\%}     & {971.9}   & {0.176}     & {19.2}   & {37.3\%}\\
Qwen2.5-VL-origin        & {10.6\%}     & {4276.2}  & {0.083}     & {25.3}   & {73.1\%}  
                         & {0.0 \%}     & {9978.3}   & {0.000}     & {31.2}   & {94.5\%}  
                         & {8.5\%}     & {7844.0}   & {0.079}     & {26.0}   & {74.6\%}\\
Qwen2.5-VL-no-thinking   & {45.8\%}     & {595.4}  & {0.446}     & \textbf{6.9}   & \textbf{0.9\%}  
                         & {10.9 \%}     & {1340.4}   & {0.105}     & \textbf{8.2}   & \textbf{0.0\%}  
                         & {27.1\%}     & {907.5}   & {0.268}     & \textbf{7.6}   & \textbf{3.4\%}\\
Qwen2.5-VL-sft-only      & {49.1\%}     & {594.2}  & {0.424}     & {10.7}   & {5.6\%}  
                         & {22.8 \%}     & {1239.7}   & {0.224}     & {19.3}   & {18.2\%}  
                         & {35.6\%}     & {839.7}   & {0.333}     & {12.1}   & {5.1\%}\\
Qwen2.5-VL-sft-rl        & {\textbf{53.2\%}}     & {\textbf{520.9}}  & {\textbf{0.513}}     & {7.9}   & {2.8\%}  
                         & {\textbf{30.9 \%}}     & {\textbf{1162.8}}   & {\textbf{0.288}}     & {13.4}   & {7.3\%}  
                         & {\textbf{42.4\%}}     & {\textbf{824.6}}   & {\textbf{0.405}}     & {9.8}   & {5.1\%}\\
\bottomrule
\end{tabular}
}
\label{tab:type_012}
\end{table*}

\vspace{-5pt}

\begin{table*}[t]
\centering
\caption{Performance comparison on Type Multi-stage, Open Door and Pick\&Drop tasks.}
\vspace{1mm}
\scriptsize
\renewcommand{\arraystretch}{1}

\resizebox{\textwidth}{!}{
\begin{tabular}{l|ccccc|ccccc|ccccc}
\toprule
\textbf{Method} 
& \multicolumn{5}{c|}{\textbf{Multi-stage}} 
& \multicolumn{5}{c|}{\textbf{Interaction - Open Door}} 
& \multicolumn{5}{c}{\textbf{Interaction - Pick and Drop}} \\
& SR $\uparrow$ & GDE $\downarrow$ & SSPL $\uparrow$ & Steps $\downarrow$ & TR $\downarrow$
& SR $\uparrow$ & GDE $\downarrow$ & SSPL $\uparrow$ & Steps $\downarrow$ & TR $\downarrow$
& SR $\uparrow$ & GDE $\downarrow$ & SSPL $\uparrow$ & Steps $\downarrow$ & TR $\downarrow$ \\
\midrule
\multicolumn{16}{c}{\textbf{In-Domain}} \\
\midrule
GPT-4o                   & {27.3\%}     & {478.9}  & {0.194}     & {23.0}   & {40.9\%}  
                         & {28.6 \%}     & {275.2}   & {0.160}     & {25.1}   & {62.6\%}  
                         & {23.5\%}     & {2124.6}   & {0.213}     & {22.8}   & {55.9\%}\\
Gemini 2.5 Pro           & {40.9\%}     & {362.5}  & {0.383}     & {19.8}   & {40.9\%}  
                         & {53.8 \%}     & {179.8}   & {0.474}     & {10.5}   & {12.1\%}  
                         & {39.8\%}     & \textbf{1615.3}   & {0.255}     & {26.2}   & {73.5\%}\\
Qwen2.5-VL-origin        & {0.0\%}     & {643.7}  & {0.000}     & {28.3}   & {86.4\%}  
                         & {9.9 \%}     & {354.34}   & {0.043}     & {29.7}   & {84.6\%}  
                         & {0.0\%}     & {8882.8}   & {0.000}     & {32.0}   & {100.0\%}\\
Qwen2.5-VL-no-thinking   & {72.7\%}     & {278.8}  & {0.708}     & \textbf{9.7}   & {4.5\%}  
                         & {68.1 \%}     & {113.2}   & {0.648}     & \textbf{7.1}  & \textbf{1.1\%}  
                         & \textbf{64.7\%}     & {3069.3}   & \textbf{0.587}     & \textbf{13.3}   & \textbf{0.0\%}\\
Qwen2.5-VL-sft-only      & \textbf{81.8\%}    & \textbf{205.2}  & {0.707}     & {13.9}   & {4.5\%}  
                         & \textbf{80.2 \%}     & \textbf{99.9}   & \textbf{0.723}     & {8.4}   & {3.3\%}  
                         & {44.1\%}     & {2645.5}   & {0.388}     & {15.2}   & {\textbf{0.0\%}}\\
Qwen2.5-VL-sft-rl        & \textbf{81.8\%}     & {283.6}  & \textbf{0.771}     & {10.6}   & \textbf{0.0\%}  
                         & {73.6 \%}     & {107.9}   & {0.687}     & {7.6}  & {\textbf{1.1\%}}  
                         & {50.0\%}     & {2665.5}   & {0.447}     & {13.4}   & \textbf{0.0\%}\\
\midrule
\multicolumn{16}{c}{\textbf{Out-of-Domain}} \\
\midrule
GPT-4o                   & {2.7\%}     & {1312.9}  & {0.027}     & {25.8}   & {64.9\%}  
                         & {16.7 \%}     & {1804.6}   & {0.069}     & {29.1}  & {83.3\%}  
                         & {29.2\%}     & {5787.5}   & {0.270}     & {25.8}   & {70.3\%}\\
Gemini 2.5 Pro           & {16.2\%}     & {1207.4}  & {0.156}     & {17.6}   & {32.4\%}  
                         & {33.3 \%}     & {430.3}   & {0.318}     & {17.8}  & {30.3\%}  
                         & {25.0\%}     & {6746.2}   & {0.192}     & {23.1}   & {54.2\%}\\
Qwen2.5-VL-origin        & {0.0\%}     & {8788.9}  & {0.000}     & {30.7}   & {94.6\%}  
                         & {7.6 \%}     & {9761.5}   & {0.037}     & {28.8}  & {83.3\%}  
                         & {0.0\%}     & {9999.0}   & {0.000}     & {32.0}   & {100.0\%}\\
Qwen2.5-VL-no-thinking   & {10.8\%}     & {1059.9}  & {0.108}     & \textbf{11.3}   & \textbf{2.7\%}  
                         & {30.3 \%}     & {316.1}   & {0.290}     & \textbf{8.3}  & \textbf{0.0\%}  
                         & {33.3\%}     & {8431.9}   & \textbf{0.457}     & \textbf{11.8}   & \textbf{0.0\%}\\
Qwen2.5-VL-sft-only      & {18.9\%}     & \textbf{1122.5}  & {0.170}     & {18.7}   & {21.6\%}  
                         & {42.4 \%}     & {283.2}   & {0.352}     & {12.9}  & {9.1\%} 
                         & {32.8\%}     & {5147.6}  & {0.319}     & {19.6}   & {8.3\%}  \\
Qwen2.5-VL-sft-rl        & \textbf{27.0\%}     & {1265.7}  & \textbf{0.253}     & {16.5}   & {16.2\%}  
                         & \textbf{45.5 \%}     & \textbf{268.9}   & \textbf{0.381}     & {11.8}  & {6.1\%} 
                         & \textbf{37.5\%}     & \textbf{4527.3}   & {0.341}     & {14.5}   & {4.2\%}\\
\bottomrule
\end{tabular}
}
\label{tab:type_345}
\end{table*}

\subsection{Experiment Setup}
\vspace{-1pt}
To evaluate the effectiveness of \embraceThreeK{} and its contribution to embodied reasoning, we conduct experiments on both in-domain and out-of-domain tasks sampled from UnrealZoo environments. Test scenarios cover six task types defined in our benchmark: \texttt{Basic}, \texttt{Exploration}, \texttt{Dynamic Spatial-Semantic}, \texttt{Multi-stage}, \texttt{Interaction - Open Door}, and \texttt{Interaction - Pick and Drop}. This diverse coverage allows us to assess model behavior under different spatial, semantic, and sequential reasoning requirements.

Each test prompt includes a structured input consisting of the task instruction, a brief description of the current scene, and a history of previously executed actions. For the visual input, we provide the first-person egocentric views at the current time step, as well as the five most recent frames and the initial frame. This limited-frame strategy strikes a balance between temporal context and computational tractability. Including the full trajectory often leads to excessive latency and model timeout. Then, we evaluate a range of models as baselines: - \textbf{GPT-4o}, \textbf{Gemini 2.5 Pro}, and the original \textbf{Qwen2.5-VL-origin} serve as zero-shot or few-shot baselines without task-specific tuning.
- \textbf{Qwen2.5-VL-sft-rl}: Our fully fine-tuned variant, which begins with SFT on \embraceThreeK{} and is further trained using reinforcement learning with trajectory-level reward shaping.
- \textbf{Qwen2.5-VL-sft-only}: A model trained only with SFT on our dataset, without additional RL optimization.
- \textbf{Qwen2.5-VL-no-thinking}: An ablated variant trained via SFT, where all chain-of-thought (\texttt{<think>}) reasoning annotations are removed from the input. This model isolates the contribution of explicit reasoning supervision to decision-making performance.

All Qwen variants in our experiments are based on the 7B architecture. All models are tested under consistent inference conditions, using the same evaluation protocol and task sets to ensure fair comparison.

\subsection{Evaluation Metrics}
\vspace{-3pt}
We evaluate agent performance mainly using the following five metrics:\\
\textbf{Success Rate (SR):} This metric measures the proportion of tasks that the agent completes successfully. A task is considered successful if the agent reaches the goal under task-specific spatial and behavioral constraints, such as reaching within 300 meters of the target and issuing a \texttt{Finish} action.

\textbf{Goal Distance Error (GDE):}
GDE quantifies the Euclidean distance (in centimeters) between the agent’s final position and the assigned target. For multi-stage tasks, GDE is computed as the sum of distances to each subgoal, with special handling to account for missing or inaccurate midway targets.

\textbf{Step-based Success weighted by Path Length (SSPL):} 
SSPL evaluates the efficiency of successful episodes. It is defined as the ratio of the optimal number of steps to the actual steps taken, weighted by success. Specifically, for each task $i$,
\[
\text{SSPL}_i = S_i \cdot \frac{s_i^{\text{opt}}}{\max(s_i, s_i^{\text{opt}})}
\]
where $S_i$ indicates task success, $s_i^{\text{opt}}$ is the length of the shortest ground-truth action sequence, and $s_i$ is the number of actions executed by the agent.

\textbf{Steps:}
This metric reports the average number of discrete actions (e.g., \texttt{MoveForward}, \texttt{TurnLeft}) taken by the agent per episode, regardless of success or failure, reflecting the behavioral cost.

\textbf{Timeout Rate (TR):}
Timeout Rate measures the proportion of episodes in which the agent exceeds a maximum step threshold (e.g., 32 steps) without completing the task. A high TR indicates frequent inefficiencies or failures to terminate appropriately.

\subsection{Experiment Analysis}
\vspace{-2pt}
We analyze the results presented in Table~\ref{tab:type_012} and Table~\ref{tab:type_345} to assess how different models perform across task types in both in-domain and out-of-domain settings.

\paragraph{Challenge difficulty.} Across all models without fine-tuning, performance remains low on exploration, spatial-relational, and multi-stage tasks. For example, the success rate (SR) of GPT-4o is only 3.6\% (Exploration), 10.2\% (Dynamic Spatial-Semantic), and 2.7\% (Multi-stage) on out-of-domain tasks. Similar patterns are observed for Gemini 2.5 Pro and Qwen2.5-VL-origin. These results confirm that \embraceThreeK{} poses substantial challenges for zero-shot models, particularly in tasks requiring long-horizon planning and egocentric spatial reasoning.

\paragraph{Supervised and RL fine-tuning improves performance.} Fine-tuning Qwen2.5-VL with our dataset leads to strong gains across all task types. In Table~\ref{tab:type_012}, the \textbf{sft-rl} variant achieves 30.9\% SR on Exploration tasks and 42.4\% SR on Spatial-Semantic tasks out-of-domain, both higher than GPT-4o and Gemini 2.5 Pro. Notably, its GDE drops from over 9978.3 to 1162.8 on Exploration, and from 7844.0 to 824.6 on Spatial-Semantic. Similar improvements are observed in Table~\ref{tab:type_345}: on Multi-stage tasks, SR improves from 0.0\% (Qwen2.5-VL-origin) to 27.0\%, and GDE reduces from 8788.9 to 1265.7. These gains show that environment-aligned supervision paired with RL reward shaping substantially improves success rates and path efficiency.

\paragraph{Reasoning annotation improves decision quality.} Comparing \textbf{sft-only} and \textbf{no-thinking} models isolates the contribution of chain-of-thought reasoning. In Dynamic Spatial-Semantic (in-domain), SR improves from 27.1\% (no-thinking) to 42.4\%, and SSPL increases from 0.268 to 0.405. Similarly, in Multi-stage tasks, the SR gap between no-thinking (10.8\%) and sft-only (18.9\%) indicates more stable sequential behavior when decision steps are paired with language rationale. These results suggest that step-wise reasoning supervision helps maintain spatial grounding and task context.

\paragraph{Generalization remains a key challenge.} The \textbf{sft-only} model performs well in-domain but shows large performance drops out-of-domain. For instance, its SR on Exploration drops from 71.4\% to 22.8\%, and on Multi-stage from 68.6\% to 35.6\%. In contrast, the RL-augmented model generalizes better: Exploration SR is 30.9\%, and Multi-stage is 42.4\%. This supports our hypothesis that trajectory-level reinforcement signals promote policy robustness in unseen scenes, where spatial layout and object configuration differ from training environments.

\section{Conclusion}
\vspace{-3pt}
This work introduces \embraceThreeK{}, a novel dataset and benchmark designed to address the limitations of current VLMs in embodied, interactive scenarios. Featuring diverse environments and multi-actions, \embraceThreeK{} fosters research in dynamic, goal-oriented contexts within open environments. High-quality CoT annotations enhance agent actions by incorporating reasoning into spatial planning. This approach bridges the gap between instructional tasks and visual inputs, enabling more robust and logical decision-making. Benchmarking experiments reveal significant challenges in spatial reasoning, long-term planning, and causal understanding, underscoring the dataset’s value in advancing embodied reasoning. Notably, fine-tuning VLMs like Qwen2.5-VL-7B with \embraceThreeK{} achieves superior performance compared to GPT-4o and Gemini 2.5 Pro. By enabling temporal generalization and integrating perception with language-guided behavior, \embraceThreeK{} establishes a foundation for developing intelligent agents capable of real-world applications.



\clearpage

{
\small
\bibliographystyle{ieee_fullname}
\bibliography{paper.bib}
}

\end{document}